\documentclass{article}

\usepackage{PRIMEarxiv}

\usepackage[utf8]{inputenc} 
\usepackage[T1]{fontenc}    
\usepackage{hyperref}       
\usepackage{url}            
\usepackage{booktabs}       
\usepackage{amsfonts}       
\usepackage{nicefrac}       
\usepackage{microtype}      
\usepackage{lipsum}
\usepackage{fancyhdr}       
\usepackage{graphicx}       
\graphicspath{{media/}}     

\usepackage{amsmath} 

\title{TGPR: Tree-Guided Policy Refinement for Robust Self-Debugging of LLMs
\thanks{\textit{\underline{Citation}}: 
\textbf{Authors. Title. Pages.... DOI:000000/11111.}} 
}
\author{
Daria Ozerova \\
HSE University, Moscow, Russia \\
\texttt{daozerova@edu.hse.ru}
\and
\textbf{Ekaterina Trofimova} \\
HSE University, Moscow, Russia \\
\texttt{etrofimova@hse.ru}
}

\begin{document}
\maketitle

\begin{abstract}
Iterative refinement has been a promising paradigm to enable large language models (LLMs) to resolve difficult reasoning and problem-solving tasks. One of the key challenges, however, is how to effectively search through the enormous search space of possible refinements. Existing methods typically fall back on predefined heuristics, which are troubled by the exploration–exploitation dilemma and cannot adapt based on past refinement outcomes. We introduce Tree-Guided Policy Refinement (TGPR), a novel framework that combines GRPO with a Thompson-Sampling-based tree search. TGPR explores both failed and successful refinement paths actively, with denser training trajectories and more adaptive policies. On HumanEval, MBPP, and APPS benchmarks, our method achieves up to +4.2 percentage points absolute improvement in pass@1 (on MBPP) and up to +12.51 percentage points absolute improvement in pass@10 (on APPS) compared to a competitive GRPO baseline. Apart from debugging code, TGPR focuses on a principled approach to combining learned policies with structured search methods, offering a general framework for enhancing iterative refinement and stateful reasoning in LLMs.
\end{abstract}


\section{Introduction}

Large Language Models (LLMs) have demonstrated remarkable capabilities in automated code generation, successfully translating natural language descriptions into syntactically correct and often functional code \cite{guo2024exploring, liu2024evaluating}. However, their performance on complex programming benchmarks like HumanEval, MBPP, and APPS reveals a critical limitation: single-pass generation often fails when faced with intricate algorithms or subtle bugs \cite{yu2024humaneval, hendrycks2021measuring}. This underscores the necessity of iterative refinement, a paradigm where models progressively debug and enhance their initial output based on execution feedback. This paradigm has become a key strategy for tackling complex, stateful reasoning tasks in code \cite{zheng2024opencodeinterpreter}.

The central challenge of iterative refinement is one of efficient exploration. The process of debugging defines a vast search tree of possible code repairs, and navigating this space effectively is paramount. Prior methods, however, often rely on fixed heuristics or standard reinforcement learning (RL) algorithms like Group Relative Policy Optimization (GRPO) that depend on the policy's own sampling for exploration \cite{shao2024deepseekmath, tang2024code}. These approaches struggle with the exploration-exploitation dilemma; they lack an adaptive mechanism to learn from the refinement process itself, often leading to myopic searches that get stuck in local optima or inefficiently explore unpromising paths \cite{jiang2024ledex}. This critical gap - the absence of a strategic exploration mechanism limits their ability to find correct solutions and learn robust debugging policies.

To address this gap, we introduce \textbf{Tree-Guided Policy Refinement (TGPR)}, a novel framework that enhances the GRPO algorithm by leveraging a Thompson Sampling-guided tree search to generate more effective and diverse training trajectories. The key differentiation in our approach is that the tree search is not used for test-time inference, but rather as a powerful data augmentation engine during training. By using a principled bandit algorithm, Thompson Sampling, to guide the exploration of the code repair tree, TGPR adaptively balances exploring uncertain edits and exploiting promising ones.

This synergy allows the agent to learn from a strategically explored set of both successful and failed refinement paths, overcoming the limitations of naive on-policy exploration. The tree search generates a high-quality, diverse dataset of trajectories that provides a much richer learning signal to the GRPO policy than it could ever discover on its own. Consequently, the policy learns a more robust and generalizable debugging strategy, internalizing the lessons from the structured search process.

Compared to a standard GRPO baseline, our tree-guided method achieves an absolute pass@1 improvement +4.2 percentage points and a pass@10 improvement of 12.5 percentage points across the HumanEval, MBPP, and APPS benchmarks. Our work presents a new, principled paradigm for automated code debugging, demonstrating that combining learned policies with structured search methods is a powerful approach for tackling complex, stateful reasoning tasks. Our contributions are:
\begin{enumerate}
    \item We introduce TGPR, a framework that enhances a strong policy gradient algorithm with a Thompson Sampling-guided tree search for strategic exploration during training.
    \item We show that using this tree search as a data generation mechanism creates more effective and diverse training trajectories, allowing the agent to learn from both successes and informative failures.
    \item We achieve the performance improvements over a strong GRPO baseline across multiple challenging code generation benchmarks.
\end{enumerate}

\section{Related Works}

Iterative refinement and debugging of LLM-generated code is a promising avenue for creating complex software, fundamentally grappling with the exploration-exploitation dilemma. The challenge lies in balancing the exploitation of partially working code variants with the exploration of less obvious correction paths. This problem has been formalized as an arm-acquiring bandit problem \cite{tang2024code}, where each "arm" represents a program variant, and "pulling an arm" corresponds to refining that program with an LLM. The goal is to solve the programming task with minimal LLM interactions.

To navigate the vast search space of code repairs, tree-based search methods have emerged as powerful tools. Algorithms like REx \cite{tang2024code} apply Bayesian approaches such as Thompson Sampling to iteratively refine programs. By maintaining probabilistic beliefs about the optimality of each program variant and updating these based on observed rewards, these methods dynamically balance exploration of uncertain paths and exploitation of promising ones. Similarly, Monte Carlo Tree Search (MCTS) has been explored for comprehensive exploration in LLM-based heuristic design \cite{zheng2025monte}, aligning with the concept of navigating a refinement tree. Our work extends these ideas by integrating a Thompson Sampling-guided tree search as a novel data augmentation engine for GRPO, specifically tailored for robust self-debugging of LLMs during training.

Beyond iterative tree-based approaches, other methods contribute to LLM-based code debugging. Some focus on static analysis and iterative generation. For instance, LDB (Language Model Debugger) \cite{zhong2024debug} employs execution profiling and step-by-step debugging with LLMs to identify and correct errors at the basic block level, mimicking human debugging processes. It analyzes runtime information to localize semantic errors more precisely. Additionally, Reinforcement Learning (RL) has been widely adopted for fine-tuning code debugging agents. Frameworks like LeDex \cite{jiang2024ledex} use a two-stage approach involving supervised fine-tuning (SFT) on "explanation + refinement" pairs, followed by RL with distinct rewards for code quality and explanation quality. Lastly, Retrieval-Augmented Generation (RAG) systems, such as CoCoGEN \cite{bi2024iterative}, enhance LLM code generation by incorporating project-level context and iterative refinement based on compiler feedback, particularly for project-scale code generation where LLMs often lack full contextual understanding.
Our work specifically builds upon the principles of iterative refinement and the exploration-exploitation balance, differentiating itself by applying a Thompson Sampling-guided tree search as a novel training-time data augmentation engine to improve the robustness of GRPO-trained LLM agents for code debugging.

\section{Approach}

\begin{figure*}[ht!]
    \includegraphics[width=1\textwidth]{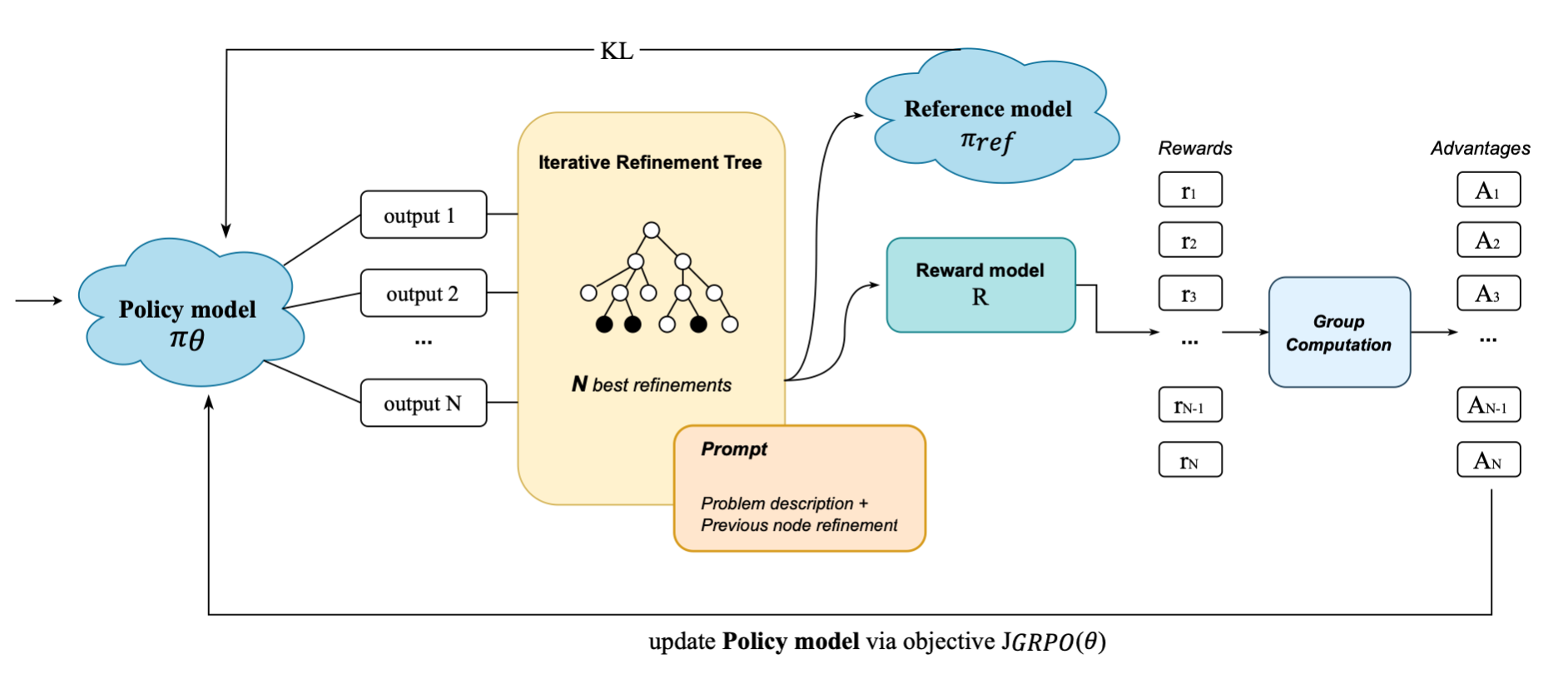}
    \caption{The overall architecture of the TGPR pipeline. The Policy Model generates candidate solutions, which populate the Iterative Refinement Tree. The tree, guided by Thompson Sampling, selects promising nodes for further refinement. The outcomes are evaluated by the Reward Model, and the resulting trajectories (states, actions, rewards, advantages) are used to update the Policy Model via the GRPO objective.}
    \label{fig:pipeline_image}
\end{figure*}

The TGPR framework is designed to overcome the exploration challenges of standard GRPO by integrating a policy model with a structured tree search. The full architecture, shown in Figure \ref{fig:pipeline_image}, illustrates how the policy model, reward model, and the Thompson Sampling-guided tree work in concert to generate high-quality training trajectories for policy refinement.

\subsection{Preliminary}

Group Relative Policy Optimization (GRPO) \cite{guo2025deepseek} is a reinforcement learning algorithm designed to optimize language models efficiently without requiring an explicit value function or critic. GRPO modifies the standard Proximal Policy Optimization (PPO) \cite{schulman2017proximal} objective by computing token-level advantages based on group-normalized rewards, making it particularly suitable for LLMs.

Given a batch of $G$ responses $\{o_i\}_{i=1}^G$ from a query $q$, each consisting of a sequence of tokens $o_i = (o_{i}(1), ..., o_{i}(T))$, the GRPO objective is defined as:
\[
    J_{\text{GRPO}}(\theta) = \frac{1}{G}\sum_{i=1}^G \frac{1}{|o_i|}\sum_{t=1}^{|o_i|} \min \left[ \frac{\pi_\theta(o_i(t)|o_{i,<t})}{\pi_{\text{old}}(o_i(t)|o_{i,<t})} \text{clip}\left( \frac{\pi_\theta(o_i(t)|o_{i,<t})}{\pi_{\text{old}}(o_i(t)|o_{i,<t})}, 1-\varepsilon, 1+\varepsilon \right) \hat{A}_{i,t} \right],
\]
where $\hat{A}_{i,t}$ is the group-normalized advantage for token $t$ in response $o_i$, computed as:
\[
    \hat{A}_{i,t} = \frac{r_i - \mu}{\sigma}, \text{ with } r_i \text{ the total reward of } o_i,
\]
and $\mu$, $\sigma$, the mean and standard deviation of rewards in the group.

The GRPO algorithm, like many policy gradient methods, fundamentally relies on the agent's ability to explore and generate diverse, high-quality trajectories during training. However, in complex, sparse-reward environments such as iterative code refinement, the policy's own exploration can be inefficient and often stagnates, limiting the discovery of optimal or even merely improved solutions. This inherent limitation in exploration efficiency forms a critical bottleneck that motivates our proposed approach.


\subsection{MDP Formulation and Custom Reward Design }
\label{sec:reward}
Unlike single-turn reinforcement learning, code refinement agents are required to perform multi-turn reasoning and decision-making, interacting with dynamic environments that provide feedback (e.g., compiler errors, test results). We adopt a Markov Decision Process (MDP) framework, where each agent trajectory comprises a sequence of code observations $s_t$, token actions $a_t$, and a scalar reward $r$ at the end of the trajectory. The agent policy $\pi_\theta$ is optimized to maximize the rewards:
\[
    \tau = \{s_t, a_t\}_{t=0,1,...,T-1}, 
    \]
    \[
    \text{ where } a_t \sim P_\theta(s_t, \{a_i\}_{i<t}).
\]
Here, a state $s \in \mathcal{S}$ represents the current code program, including an initial faulty version, compiler errors, test case failures, or any other relevant feedback. An action $a \in \mathcal{A}$ is a modification to the current code program, generated as a sequence of tokens.

A critical component of our system is a well-defined reward function $R(\rho)$ that provides a dense and informative signal to both the RL agent and the tree search. A sparse binary reward (i.e., 1 for passing all tests, 0 otherwise), as is common in the MDP formulation, is insufficient for guiding the nuanced process of debugging. Therefore, we design a hybrid reward function that combines signals of functional correctness and semantic similarity. The reward for a given program refinement $\rho$ is defined as:
$$ R(\rho) = \text{CodeBLEU}(\rho, \rho_c) + \frac{|T_p(\rho)|}{|T|} $$
where:
\begin{itemize}
    \item $\rho_c$ is the canonical reference solution from the benchmark.
    \item $\text{CodeBLEU}(\rho, \rho_c)$ measures the syntactic and semantic similarity between the generated code $\rho$ and the reference solution $\rho_c$. This provides a continuous signal of progress even if the code is not yet functionally correct.
    \item $|T_p(\rho)|$ is the number of unit tests successfully passed by the program $\rho$.
    \item $|T|$ is the total number of unit tests for the given problem. This component directly measures the functional correctness of the code.
\end{itemize}
This composite reward function is used consistently across the entire framework, providing a unified quality metric for training the reward model, guiding the tree search, and calculating advantages for the GRPO update. This reward is predicted by a specially trained reward model $r_\phi$.

\subsection{Thompson Sampling-Guided Tree Search}

To address the exploration bottleneck of standard RL and organize the iterative debugging process, we employ a refinement tree guided by Thompson Sampling. This tree is integrated into the GRPO architecture and serves as an effective mechanism for searching and selecting the most promising code refinement candidates generated by the language model.

Each node in this tree corresponds to a specific version of the program (refinement), and edges represent refinement actions. During the iterative process, multiple new code variants are generated at each step and added to the tree as children of the current node. The selection of the next program to refine balances the exploration of new variants with the exploitation of partially successful solutions.

For selecting the next node in the tree to refine, we employ a stochastic strategy based on Thompson Sampling with a Beta distribution. For each program $\rho$ (node) in the tree, we compute parameters $\alpha_\rho$ and $\beta_\rho$:
\begin{equation}
\alpha_\rho = 1 + C \cdot R(\rho)
\end{equation}
\begin{equation}
\beta_\rho = 1 + C \cdot (1 - R(\rho)) + N_\rho
\end{equation}
where:
\begin{itemize}
    \item $C$ is a hyperparameter that controls the degree of confidence (and thus the balance between exploration and exploitation).
    \item $N_\rho$ is the number of unsuccessful refinement attempts originating from program $\rho$.
    \item $R(\rho)$ is the unified reward for program $\rho$, as defined in the previous subsection.
\end{itemize}
Subsequently, for each program $\rho$, a value $\theta_\rho^{(s)}$ is sampled from a Beta distribution: $\theta_\rho^{(s)} \sim \text{Beta}(\alpha_\rho, \beta_\rho)$. The program with the maximum sampled value is then chosen as the next program for refinement:
\begin{equation}
\rho_{\text{next}} = \arg\max_\rho \theta_\rho^{(s)}
\end{equation}
This Thompson Sampling-guided selection strategy allows for dynamic balancing between exploring less-visited but potentially high-reward paths and exploiting known good solutions. The tree thus provides a structured way to manage the exploration of the immense search space of code refinements, contributing higher-quality and more diverse training trajectories to the GRPO agent.

Notably, tree search is applied only at training. Applying it at inference would be too computationally expensive and slow because every refinement step would require checking a few candidate programs. Restricting tree search to training, the GRPO agent learns the strategic exploration patterns discovered using Thompson Sampling, learning how to balance exploration and exploitation and make a single-shot, confident decisions during test time. This technique uses the tree as a data augmentation and exploration plan of high dimensionality, enabling policy learning without experiencing inference-time costs.

\begin{table*}[bt!]
    \centering
    \caption{Summary of LLM-generated code samples and verified refinements for each benchmark dataset.}
    \label{tab:unique_correct_wrong_solutions}
    \begin{tabular}{lrrrrr}
        \toprule
        \textbf{Dataset} & \textbf{\#Unique Sol.} & \textbf{\#Correct Sol.} & \textbf{\#Wrong Sol.} & \textbf{\#Correct Ref.} & \textbf{\#Ref. Rate} \\
        \midrule
        MBPP Training & 9,500 & 4,706 & 4,794 & \textbf{2,450} & \textbf{51.10}\% \\ 
        HumanEval Training & 44,108 & 27,736 & 16,372 & \textbf{7,100} & \textbf{43.37}\% \\ 
        APPS Training & 51,134 & 31,520 & 19,614 & \textbf{6,000} & \textbf{30.59}\% \\ 
        \bottomrule
    \end{tabular}
\end{table*}

\section{Experiments}

\subsection{Implementation Details}
\label{sec:implementation_details}

\paragraph{Policy Model}
 We use a fine-tuned Qwen-7B model as the base policy model, which serves as our agent for code refinement. Training is conducted using a reinforcement learning realization from the Hugging Face ecosystem. For trajectory rollout, we set up 256 parallel virtual environments, and the rollout number for each task is 8. A total of 128 tasks are sampled from a combined set of MBPP's training set and CodeContests, according to the strategy described in Sec. 4.2, and training is performed over 5 epochs. Rollouts are conducted with a batch size of 32 and a temperature of 1.0 to encourage exploration during data generation for GRPO. For policy optimization, we use the AdamW optimizer \cite{loshchilov2017decoupled} with a learning rate of $1 \times 10^{-6}$ and a mini-batch size of 8 per device. The gradient accumulation number is 4. Following DAPO \cite{yu2025dapo}, we set the clipping parameters for the GRPO objective to $\epsilon_{\text{low}} = 0.2$ and $\epsilon_{\text{high}} = 0.3$ to balance exploration and exploitation. During evaluation, the temperature is lowered to 0.6 for more stable and deterministic performance. We explicitly remove the KL divergence loss term from the objective to remove the need for a separate reference model, simplifying the training pipeline.

\paragraph{Reward Model}
The reward model $r_{\phi}$ is the fine-tuned Qwen-7B base as the policy model.

Training runs for 3 epochs using AdamW (learning rate $5\times10^{-6}$, batch size 16), 
with an 80/10/10 split stratified by reward value. 
Dropout ($p=0.1$) and early stopping on validation MSE prevent overfitting.

\paragraph{Integration with GRPO}
The trained $r_{\phi}$ evaluates all generated samples during GRPO training. 
Optional replay-based fine-tuning provides marginal gains (+0.3~pp pass@1) and is omitted for simplicity. 
On held-out data, $r_{\phi}$ achieves strong correlation with ground-truth rewards (Pearson $r>0.85$), 
confirming its reliability as a unified quality metric for both GRPO optimization and TGPR tree navigation.

 \begin{figure*}[]
    \centering
    \includegraphics[width=0.75\textwidth]{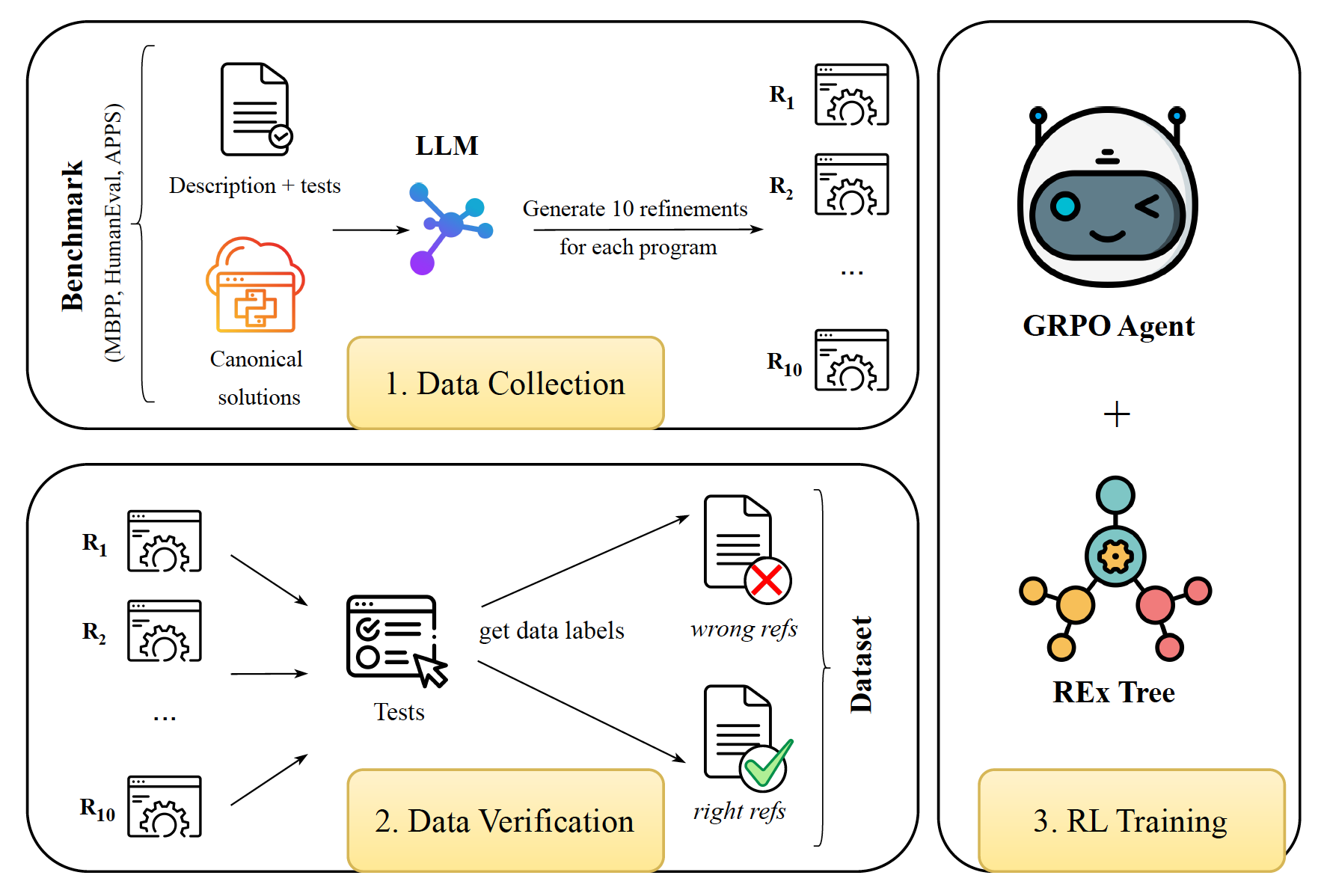}
    \caption{Data Preprocessing Pipeline for Code Debugging Module.}
\end{figure*}

 All experiments were conducted on a single server equipped with A100 GPUs (80GB VRAM each), providing ample computational resources for training and running the Qwen-7B policy model and the parallel environments. Appendix~\label{sec:appendix} show the dinamic quality and stability of custom reward.

\subsection{Data collection and verification}
\label{sec:data_collection_verification}

We utilize established benchmarks: MBPP \cite{austin2021program}, APPS \cite{hendrycks2021measuring}, and CodeContests \cite{wang2025codecontests+}, all available via Hugging Face. While HumanEval \cite{chen2021evaluating} is a common evaluation standard, MBPP and APPS offer more comprehensive training and test examples. Our training data specifically comprises the MBPP training set and CodeContests, with final models evaluated on MBPP, HumanEval, and APPS.

Our data preprocessing involved prompting an LLM (GPT-3.5-turbo) for each benchmark task (with description, reference solution, and tests) to generate ten diverse code "refinements." This aimed to create a representative sample of LLM-generated erroneous code. Each generated variant was then automatically tested and classified as:
\begin{itemize}
    \item\textbf{Fully Correct:} Passed all tests.
    \item\textbf{Partially Correct:} Passed some tests, indicating residual errors.
    \item\textbf{Incorrect:} Failed all tests or resulted in runtime errors.
\end{itemize}
This process yielded structured, labeled data for system development.

To further augment our dataset with wrong solutions and their refinements, we used a 3-shot prompting strategy with pre-trained LLMs to sample 20 solutions per problem from the MBPP, APPS, and CodeContests training sets. Solutions failing any test cases were identified as "wrong." For each wrong solution, we then prompted LLMs with the problem, wrong solution, and execution feedback to generate an explanation and a *correct* refinement. This collection process is vital for creating high-quality paired examples of flawed code and debugging steps.
\begin{table*}[h!]
\centering
\caption{Pass@k on three benchmarks for Qwen-7B. Best performance bolded.}
\label{tab:results}
\resizebox{0.65\textwidth}{!}{%
\begin{tabular}{@{}lcccccc@{}}
\toprule
\textbf{Approach} & \multicolumn{2}{c}{\textbf{MBPP}} & \multicolumn{2}{c}{\textbf{HumanEval}} & \multicolumn{2}{c}{\textbf{APPS}} \\
\cmidrule(lr){2-3} \cmidrule(lr){4-5} \cmidrule(lr){6-7}
 & pass@1 & pass@10 & pass@1 & pass@10 & pass@1 & pass@10 \\
\midrule
Pretrained & 22.5 & 45.8 & 18.7 & 38.2 & 12.3 & 28.9 \\
GRPO & 26.8 & 52.1 & 22.4 & 43.6 & 15.1 & 34.2 \\
PPO & 25.3 & 49.7 & 21.6 & 41.8 & 14.7 & 32.9 \\
TGPR & \textbf{31.0} & \textbf{56.3} & \textbf{25.1} & \textbf{49.8} & \textbf{18.9} & \textbf{46.7} \\
\bottomrule
\end{tabular}%
}
\end{table*}

Table~\ref{tab:unique_correct_wrong_solutions} summarizes the number of unique, correct, and wrong solutions sampled, along with the correct refinements generated by GPT-3.5-Turbo and their respective refinement rates for each dataset.

Critically, LLM-generated explanations or refinements are not blindly trusted. We verify all refinements by re-running them against test cases, considering only those passing all tests as genuinely correct. This rigorous verification ensures the high quality of our automatically collected code explanation and refinement dataset.
\subsection{Experimental Results}
\begin{figure*}[bt!]
    \centering
    \includegraphics[width=0.7\linewidth, trim={0 0 0 10}, clip]{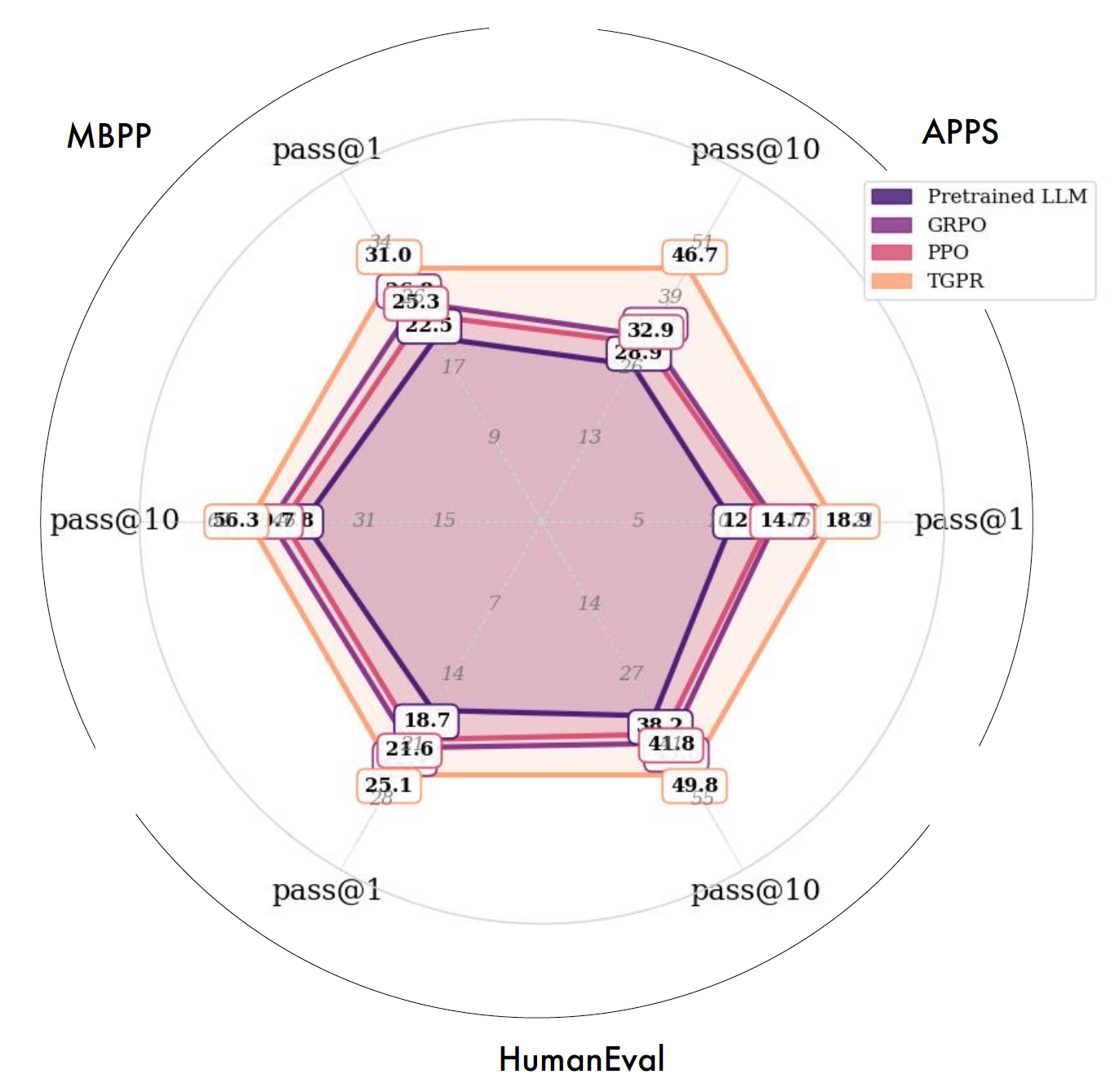}
    \caption{Passing rate analytics for Qwen-7B.}
    \label{fig:results}
\end{figure*}

\begin{figure*}[t!]
    \centering
    \includegraphics[width=0.8\linewidth, trim={0 30 0 10}, clip]{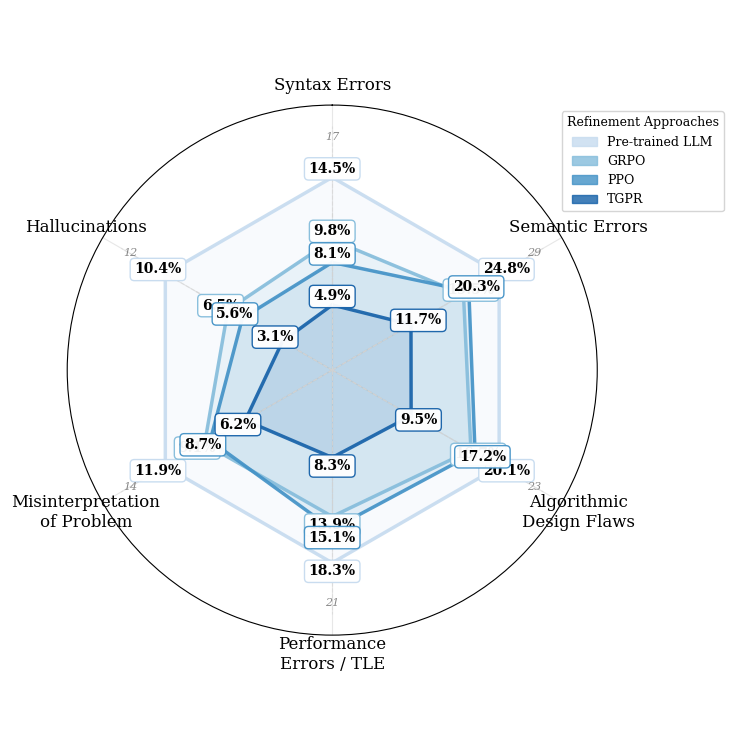}
    \caption{Error classification in code debugging.}
    \label{fig:error_rate}
\end{figure*}

\subsubsection{Overall Performance Improvements}
Table \ref{tab:results} summarizes the overall pass@k results on MBPP, HumanEval, and APPS benchmarks, detailing the performance of the Qwen-7B model with various refinement approaches.

As evident from Table \ref{tab:results}, Qwen-7B with TGPR consistently achieves the highest pass@1 and pass@10 scores across all three benchmarks, with all six performance metrics bolded. TGPR demonstrates significant improvements over both the base pretrained LLM and traditional reinforcement learning baselines (GRPO and PPO), establishing its superior capabilities in iterative code refinement.

For a direct comparison against the competitive GRPO baseline, TGPR exhibits notable performance enhancements across all benchmarks:
\begin{itemize}
\item On \textbf{MBPP}, TGPR boosts pass@1 by \textbf{4.2 percentage points} (from 26.8\% to 31.0\%) and pass@10 by \textbf{4.2 percentage points} (from 52.1\% to 56.3\%).
\item On \textbf{HumanEval}, TGPR improves pass@1 by \textbf{2.7 percentage points} (from 22.4\% to 25.1\%) and pass@10 by \textbf{6.2 percentage points} (from 43.6\% to 49.8\%).
\item On \textbf{APPS}, which features more rigorous and complex test cases, TGPR achieves a pass@1 improvement of \textbf{3.8 percentage points} (from 15.1\% to 18.9\%) and the highest overall pass@10 gain of \textbf{12.51 percentage points} (from 34.2\% to 46.71\%).
\end{itemize}
These results collectively demonstrate that TGPR training enables LLMs to produce or refine solutions that are markedly more robust and capable of passing stricter test cases. The strategic exploration guided by Thompson Sampling allows the model to learn from a more diverse and informative set of successful and failed refinement paths, leading to the development of a more resilient and effective debugging strategy.

\subsection{Error Analysis of Refinement Approaches}

This section analyzes the distribution of error categories for each code refinement approach, as illustrated in Figure \ref{fig:error_rate}. A nuanced understanding of these categories is pivotal for assessing the efficacy and limitations of advanced code generation and debugging paradigms.

\subsubsection{Error Taxonomy}
\begin{itemize}
    \item \textbf{Syntax Errors:} Fundamental language construct violations or trivial logical inconsistencies.
    \item \textbf{Semantic Errors:} Syntactically valid code yielding incorrect results under specific, often boundary or degenerate, input conditions.
    \item \textbf{Algorithmic Design Flaws:} Misapplication of algorithmic paradigms or selection of suboptimal data structures, failing to meet problem constraints (e.g., complexity).
    \item \textbf{Performance Errors / TLE:} Functionally correct code failing due to computational inefficiency (e.g., excessive time or memory consumption) on large inputs.
    \item \textbf{Misinterpretation of Problem Statement:} Generation of code addressing a tangentially related problem due to misconstrued problem specifications or implicit constraints.
    \item \textbf{Hallucinations:} Output resembling code but containing non-functional constructs, references to non-existent entities, or unresolved placeholders.
\end{itemize}

\subsubsection{Comparative Analysis}
The baseline Pre-trained LLM (Qwen 7B) exhibits the highest error rates across all categories, particularly in \textit{Semantic Errors} (24.8\%) and \textit{Algorithmic Design Flaws} (20.1\%), underscoring its foundational limitations without iterative correction. Both GRPO and PPO, as reinforcement learning-based refinement approaches, reduce these error rates. GRPO demonstrates strong improvements in \textit{Syntax Errors} (9.8\%) and \textit{Hallucinations} (6.5\%), and notable gains in \textit{Semantic Errors} (19.5\%) and \textit{Algorithmic Design Flaws} (16.7\%). PPO marginally outperforms GRPO in \textit{Syntax Errors} (8.1\%) and \textit{Hallucinations} (5.6\%), yet shows slightly higher error rates than GRPO in more complex categories like \textit{Semantic Errors} (20.3\%) and \textit{Algorithmic Design Flaws} (17.2\%), suggesting potential trade-offs in its exploration-exploitation balance for intricate logical debugging.

Crucially, our proposed TGPR approach consistently achieves the lowest error rates across all categories. This includes remarkable reductions in \textit{Algorithmic Design Flaws} (9.5\%), \textit{Semantic Errors} (11.7\%), and \textit{Performance Errors / TLE} (8.3\%). This superior performance is attributed to TGPR's Thompson Sampling-guided tree search, which facilitates strategic exploration of diverse refinement trajectories during training. This methodology enables the learning of a more robust and generalized debugging policy, mitigating even subtle errors. In conclusion, while iterative refinement generally enhances code quality, TGPR's principled integration of strategic search within the training regimen yields a demonstrably more effective and resilient debugging agent, particularly for complex algorithmic and semantic challenges.

\section{Conclusion}
This work underscores the critical importance of training open-source LLMs for robust self-debugging and introduces TGPR, a scalable framework that enhances LLMs' iterative code refinement capabilities. TGPR integrates a Thompson Sampling-guided tree search as a novel training-time data augmentation engine, coupled with a custom hybrid reward design within a GRPO-based reinforcement learning pipeline. This approach strategically generates high-quality and diverse training trajectories, enabling LLMs to develop a more effective and resilient debugging strategy. Our evaluations on MBPP, HumanEval, and APPS benchmarks demonstrate substantial performance gains from the TGPR approach. Compared to the base pretrained Qwen-7B model, TGPR achieved a significant absolute improvement in pass@1 of 8.5 percentage points (on MBPP, from 22.5\% to 31.0\%) and a notable absolute improvement in pass@10 of 17.81 percentage points (on APPS, from 28.9\% to 46.71\%). When compared against the strongest reinforcement learning baseline, GRPO, TGPR consistently delivered superior results, with absolute pass@1 improvements of up to 4.2 percentage points (on MBPP) and pass@10 improvements of up to 12.51 percentage points (on APPS). These substantial and consistent improvements across all benchmarks and metrics underscore the effectiveness of our method over both the pretrained model and traditional RL fine-tuning approaches. These results confirm the efficacy and generalizability of our framework in boosting iterative code refinement and debugging, paving the way for more autonomous and proficient LLM-powered code generation.

\bibliographystyle{unsrtnat}
\bibliography{references}

\begin{thebibliography}{18}
\providecommand{\natexlab}[1]{#1}
\providecommand{\url}[1]{\texttt{#1}}
\expandafter\ifx\csname urlstyle\endcsname\relax
  \providecommand{\doi}[1]{doi: #1}\else
  \providecommand{\doi}{doi: \begingroup \urlstyle{rm}\Url}\fi

\bibitem[Guo et~al.(2024)Guo, Cao, Xie, Liu, Li, Chen, and Peng]{guo2024exploring}
Qi~Guo, Junming Cao, Xiaofei Xie, Shangqing Liu, Xiaohong Li, Bihuan Chen, and Xin Peng.
\newblock Exploring the potential of chatgpt in automated code refinement: An empirical study.
\newblock In \emph{Proceedings of the 46th IEEE/ACM International Conference on Software Engineering}, pages 1--13, 2024.

\bibitem[Liu et~al.(2024)Liu, Xie, Wang, Wei, Ding, and Zhang]{liu2024evaluating}
Jiawei Liu, Songrun Xie, Junhao Wang, Yuxiang Wei, Yifeng Ding, and Lingming Zhang.
\newblock Evaluating language models for efficient code generation.
\newblock \emph{arXiv preprint arXiv:2408.06450}, 2024.

\bibitem[Yu et~al.(2024)Yu, Zhao, Cohan, and Zhang]{yu2024humaneval}
Zhaojian Yu, Yilun Zhao, Arman Cohan, and Xiao-Ping Zhang.
\newblock Humaneval pro and mbpp pro: Evaluating large language models on self-invoking code generation.
\newblock \emph{arXiv preprint arXiv:2412.21199}, 2024.

\bibitem[Hendrycks et~al.(2021)Hendrycks, Basart, Kadavath, Mazeika, Arora, Guo, Burns, Puranik, He, Song, et~al.]{hendrycks2021measuring}
Dan Hendrycks, Steven Basart, Saurav Kadavath, Mantas Mazeika, Akul Arora, Ethan Guo, Collin Burns, Samir Puranik, Horace He, Dawn Song, et~al.
\newblock Measuring coding challenge competence with apps.
\newblock \emph{arXiv preprint arXiv:2105.09938}, 2021.

\bibitem[Zheng et~al.(2024)Zheng, Zhang, Shen, Liu, Lin, Fu, Chen, and Yue]{zheng2024opencodeinterpreter}
Tianyu Zheng, Ge~Zhang, Tianhao Shen, Xueling Liu, Bill~Yuchen Lin, Jie Fu, Wenhu Chen, and Xiang Yue.
\newblock Opencodeinterpreter: Integrating code generation with execution and refinement.
\newblock \emph{arXiv preprint arXiv:2402.14658}, 2024.

\bibitem[Shao et~al.(2024)Shao, Wang, Zhu, Xu, Song, Bi, Zhang, Zhang, Li, Wu, et~al.]{shao2024deepseekmath}
Zhihong Shao, Peiyi Wang, Qihao Zhu, Runxin Xu, Junxiao Song, Xiao Bi, Haowei Zhang, Mingchuan Zhang, YK~Li, Y~Wu, et~al.
\newblock Deepseekmath: Pushing the limits of mathematical reasoning in open language models.
\newblock \emph{arXiv preprint arXiv:2402.03300}, 2024.

\bibitem[Tang et~al.(2024)Tang, Hu, Zhou, Zhong, Zheng, Si, and Ellis]{tang2024code}
Hao Tang, Keya Hu, Jin Zhou, Si~Cheng Zhong, Wei-Long Zheng, Xujie Si, and Kevin Ellis.
\newblock Code repair with llms gives an exploration-exploitation tradeoff.
\newblock \emph{Advances in Neural Information Processing Systems}, 37:\penalty0 117954--117996, 2024.

\bibitem[Jiang et~al.(2024)Jiang, Li, Wang, Zhou, Hossain, Ray, Kumar, Ma, and Deoras]{jiang2024ledex}
Nan Jiang, Xiaopeng Li, Shiqi Wang, Qiang Zhou, Soneya Hossain, Baishakhi Ray, Varun Kumar, Xiaofei Ma, and Anoop Deoras.
\newblock Ledex: Training llms to better self-debug and explain code.
\newblock \emph{Advances in Neural Information Processing Systems}, 37:\penalty0 35517--35543, 2024.

\bibitem[Zheng et~al.(2025)Zheng, Xie, Wang, and Hooi]{zheng2025monte}
Zhi Zheng, Zhuoliang Xie, Zhenkun Wang, and Bryan Hooi.
\newblock Monte carlo tree search for comprehensive exploration in llm-based automatic heuristic design.
\newblock \emph{arXiv preprint arXiv:2501.08603}, 2025.

\bibitem[Zhong et~al.(2024)Zhong, Wang, and Shang]{zhong2024debug}
Li~Zhong, Zilong Wang, and Jingbo Shang.
\newblock Debug like a human: A large language model debugger via verifying runtime execution step-by-step.
\newblock \emph{arXiv preprint arXiv:2402.16906}, 2024.

\bibitem[Bi et~al.(2024)Bi, Wan, Wang, Zhang, Guan, Lu, Zhang, Sui, Jin, and Shi]{bi2024iterative}
Zhangqian Bi, Yao Wan, Zheng Wang, Hongyu Zhang, Batu Guan, Fangxin Lu, Zili Zhang, Yulei Sui, Hai Jin, and Xuanhua Shi.
\newblock Iterative refinement of project-level code context for precise code generation with compiler feedback.
\newblock \emph{arXiv preprint arXiv:2403.16792}, 2024.

\bibitem[Guo et~al.(2025)Guo, Yang, Zhang, Song, Zhang, Xu, Zhu, Ma, Wang, Bi, et~al.]{guo2025deepseek}
Daya Guo, Dejian Yang, Haowei Zhang, Junxiao Song, Ruoyu Zhang, Runxin Xu, Qihao Zhu, Shirong Ma, Peiyi Wang, Xiao Bi, et~al.
\newblock Deepseek-r1: Incentivizing reasoning capability in llms via reinforcement learning.
\newblock \emph{arXiv preprint arXiv:2501.12948}, 2025.

\bibitem[Schulman et~al.(2017)Schulman, Wolski, Dhariwal, Radford, and Klimov]{schulman2017proximal}
John Schulman, Filip Wolski, Prafulla Dhariwal, Alec Radford, and Oleg Klimov.
\newblock Proximal policy optimization algorithms.
\newblock \emph{arXiv preprint arXiv:1707.06347}, 2017.

\bibitem[Loshchilov and Hutter(2017)]{loshchilov2017decoupled}
Ilya Loshchilov and Frank Hutter.
\newblock Decoupled weight decay regularization.
\newblock \emph{arXiv preprint arXiv:1711.05101}, 2017.

\bibitem[Yu et~al.(2025)Yu, Zhang, Zhu, Yuan, Zuo, Yue, Fan, Liu, Liu, Liu, et~al.]{yu2025dapo}
Qiying Yu, Zheng Zhang, Ruofei Zhu, Yufeng Yuan, Xiaochen Zuo, Yu~Yue, Tiantian Fan, Gaohong Liu, Lingjun Liu, Xin Liu, et~al.
\newblock Dapo: An open-source llm reinforcement learning system at scale, 2025.
\newblock \emph{URL https://arxiv. org/abs/2503.14476}, 2025.

\bibitem[Austin et~al.(2021)Austin, Odena, Nye, Bosma, Michalewski, Dohan, Jiang, Cai, Terry, Le, et~al.]{austin2021program}
Jacob Austin, Augustus Odena, Maxwell Nye, Maarten Bosma, Henryk Michalewski, David Dohan, Ellen Jiang, Carrie Cai, Michael Terry, Quoc Le, et~al.
\newblock Program synthesis with large language models.
\newblock \emph{arXiv preprint arXiv:2108.07732}, 2021.

\bibitem[Wang et~al.(2025)Wang, Liu, Sun, Li, and Shen]{wang2025codecontests+}
Zihan Wang, Siyao Liu, Yang Sun, Hongyan Li, and Kai Shen.
\newblock Codecontests+: High-quality test case generation for competitive programming.
\newblock \emph{arXiv preprint arXiv:2506.05817}, 2025.

\bibitem[Chen et~al.(2021)Chen, Tworek, Jun, Yuan, Pinto, Kaplan, Edwards, Burda, Joseph, Brockman, et~al.]{chen2021evaluating}
Mark Chen, Jerry Tworek, Heewoo Jun, Qiming Yuan, Henrique Ponde De~Oliveira Pinto, Jared Kaplan, Harri Edwards, Yuri Burda, Nicholas Joseph, Greg Brockman, et~al.
\newblock Evaluating large language models trained on code.
\newblock \emph{arXiv preprint arXiv:2107.03374}, 2021.

\end{thebibliography}

\appendix

\section{Appendix}
\label{sec:appendix}

\begin{figure*}[bt!]
    \centering
    \includegraphics[width=0.8\textwidth, trim=0 0 0 20, clip]{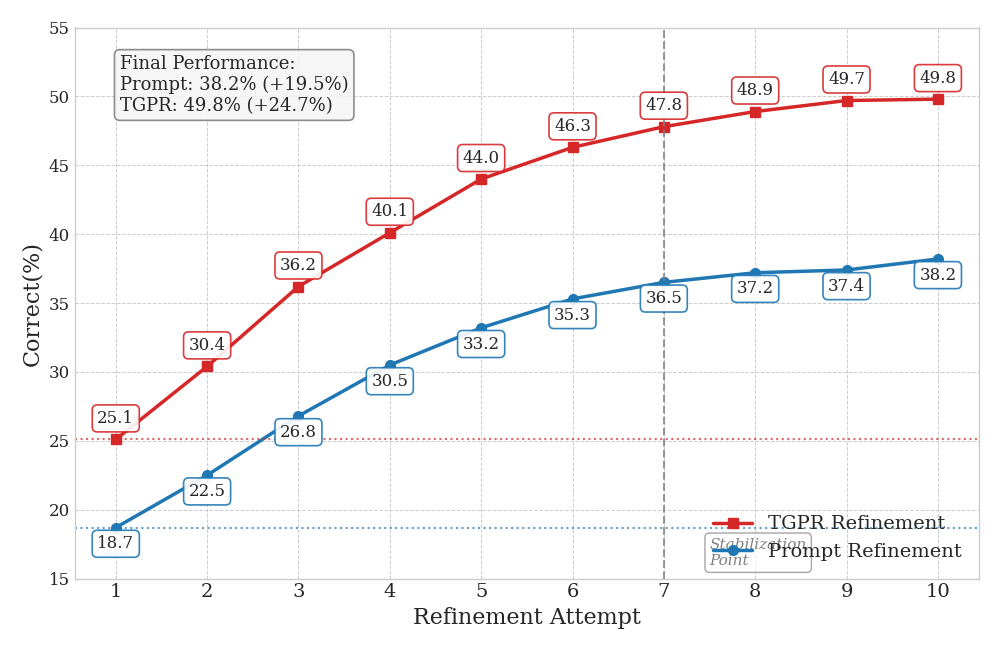}
    \caption{Iterative Refinement Quality.}
    \label{A:1}
\end{figure*}

Figure~\ref{A:1} illustrates the improvement in solution quality across refinement iterations for TGPR compared to baseline prompting. The x-axis represents the refinement iteration number, while the y-axis shows the cumulative pass rate. TGPR demonstrates steeper learning curves and higher final performance, indicating more effective utilization of each refinement step. The strategic exploration enabled by Thompson Sampling allows TGPR to make more meaningful improvements in early iterations and maintain steady progress throughout the refinement process.

\begin{figure*}[bt!]
    \centering
    \includegraphics[width=\textwidth, trim=0 0 0 0, clip]{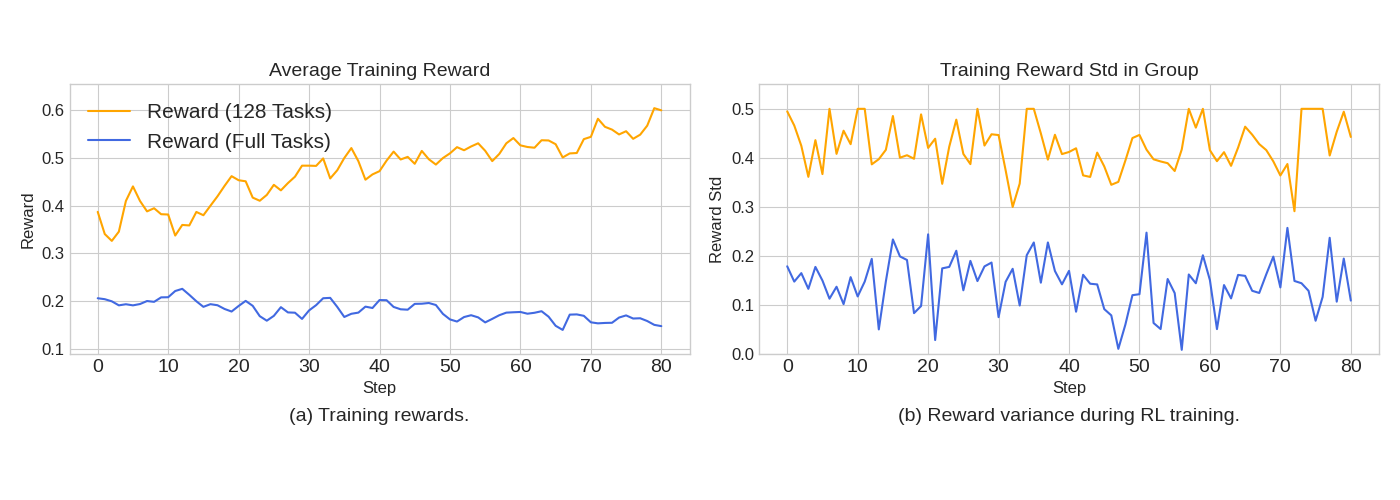}
    \caption{Reward function Stability.}
    \label{A:2}
\end{figure*}

Figure~\ref{A:2} compares the RL training performance between a selected 128-task subset and the full task set, measured using our custom reward function. The x-axis shows training steps (or episodes), and the y-axis displays the mean value of our custom reward. The selected subset achieves significantly higher average rewards while maintaining comparable reward variance to full-set training. This demonstrates that our strategic task selection improves training efficiency without increasing instability, validating the subset selection approach for more effective RL fine-tuning. Monitoring this custom reward is crucial as it provides a denser, more nuanced learning signal than sparse pass/fail metrics, incorporating code quality and syntactic correctness. The stable convergence indicates the policy is learning coherent refinement strategies from high-quality, tree-generated trajectories.

\end{document}